\documentclass{ifacconf}

\usepackage{graphicx}      
\usepackage{natbib}        

\usepackage{amsmath,amssymb,amsfonts}
\usepackage{algpseudocode,algorithm}
\usepackage{url}

\begin{document}
\begin{frontmatter}

\title{Mirror-Descent Inverse Kinematics with Box-constrained Joint Space\thanksref{footnoteinfo}}

\thanks[footnoteinfo]{This work was supported by JSPS KAKENHI, Grant-in-Aid for Scientific Research (B), Grant Number JP20H04265.}

\author[First]{Taisuke Kobayashi}
\author[Second]{Takanori Jin}

\address[First]{National Institute of Informatics (NII) and The Graduate University for Advanced Studies (SOKENDAI), 2-1-2 Hitotsubashi, Chiyoda-ku, Tokyo, 101-8430, Japan (e-mail: kobayashi@nii.ac.jp).}
\address[Second]{Division of Information Science, Nara Institute of Science and Technology, 8916-5 Takayama-cho, Ikoma, Nara 630-0192, Japan (e-mail: jin.takanori.ju0@is.naist.jp)}

\begin{abstract}                
To control humanoid robots, the reference pose of end effector(s) is planned in task space, then mapped into the reference joints by IK.
By viewing that problem as approximate quadratic programming (QP), recent QP solvers can be applied to solve it precisely, but iterative numerical IK solvers based on Jacobian are still in high demand due to their low computational cost.
However, the conventional Jacobian-based IK usually clamps the obtained joints during iteration according to the constraints in practice, causing numerical instability due to non-smoothed objective function.
To alleviate the clamping problem, this study explicitly considers the joint constraints, especially the box constraints in this paper, inside the new IK solver.
Specifically, instead of clamping, a mirror descent (MD) method with box-constrained real joint space and no-constrained mirror space is integrated with the Jacobian-based IK, so-called MD-IK.
In addition, to escape local optima nearly on the boundaries of constraints, a heuristic technique, called $\epsilon$-clamping, is implemented as margin in software level.
Finally, to increase convergence speed, the acceleration method for MD is integrated assuming continuity of solutions at each time.
As a result, the accelerated MD-IK achieved more stable and enough fast tracking performance compared to the conventional IK solvers.
The low computational cost of the proposed method mitigated the time delay until the solution is obtained in real-time humanoid gait control, achieving a more stable gait.
\end{abstract}


\end{frontmatter}

\section{Introduction}

Humanoid robots have long been studied because they can be introduced without changing human-living environments and are expected to be highly versatile~\citep{asfour2018armar,kobayashi2022whole}.
To accomplish a given task by a humanoid robot, it plans the intuitive reference poses in task space (i.e. Cartesian space for its end effector(s)), not in joint space, which is directly controllable.
In this major approach, for example, the robot can easily plan the reference poses to avoid obstacles~\citep{gienger2008task,nguyen2018compact}.

From the reference poses, the reference joints are computed basically through inverse kinematics (IK)~\citep{buss2004introduction,aristidou2018inverse}.
Only in limited cases, closed-form solutions of IK can be obtained.
Especially in humanoid robots, however, the joints activated in IK is task-dependent, so it is infeasible to prepare all the closed-form solutions for all the combinations of the activated joints in advance.
By approximating IK as a quadratic problem (QP), recent QP solvers can be applied to solve it precisely~\citep{ferreau2014qpoases,stellato2020osqp}.
However, although QP solvers are getting faster, it would be hard to solve them within the control period for the humanoid robot (roughly 1--5~ms), and even if they are solved, the delay caused by the optimization time would cause unstable motion.

Therefore, iterative numerical IK solvers with Jacobian between the task and joint spaces~\citep{sugihara2011solvability,drexler2012joint} are still popular in practice.
However, since it is a kind of gradient descent method, the computed reference joints would be in the whole real space, not in the joint space.
The joint space generally has constraints, i.e. upper and lower limits for each joint and maximum (angular) velocity for each, hence the constraints should be taken into account every iteration of Jacobian-based IK.
In practice, such constraints are satisfied with clamping the reference joints.
The numerical instability has been, however, reported due to non-smoothed objective function by clamping operation~\citep{beeson2015trac}.

To avoid this clamping problem, TRAC-IK~\citep{beeson2015trac} has proposed the use of nonlinear optimizer, which is robust to non-smooth objective function.
TRAC-IK has been established as one of the popular IK solvers, but it still remain the clamping problem.
In addition, the nonlinear optimization is costly expensive.
In contrast, Drexler and Harmati~\citep{drexler2012joint} introduces additional non-constrained variables, which are nonlinearly mapped from the box-constrained joint space and are optimized by Jacobian-based updates.
After optimization of that variables, the reference joints can be obtained by inversely mapping it without any clamping.
However, this approach has slow convergence due to the gradients added by nonlinear mapping.

In machine learning fields, the clamping operation and the above mapping technique can be regarded as a projected gradient (PG) method~\citep{calamai1987projected} and a parameterization method, which is frequently used in neural networks, respectively.
From this perspective, mirror descent (MD) method~\citep{beck2003mirror,srebro2011universality,krichene2015accelerated} can naturally emerge as another solution.
MD is a kind of generalized version of gradient descent for Bregman divergence minimization.
In MD, the constraints imposed by Bregman divergence can be explicitly considered.
In practice, thanks to this feature, MD can handle the constrained space as like the parameterization method~\citep{drexler2012joint}, while maintaining the similar convergence speed to the PG method~\citep{sugihara2011solvability}.

\begin{figure}[tb]
    \centering
    \includegraphics[keepaspectratio=true,width=0.96\linewidth]{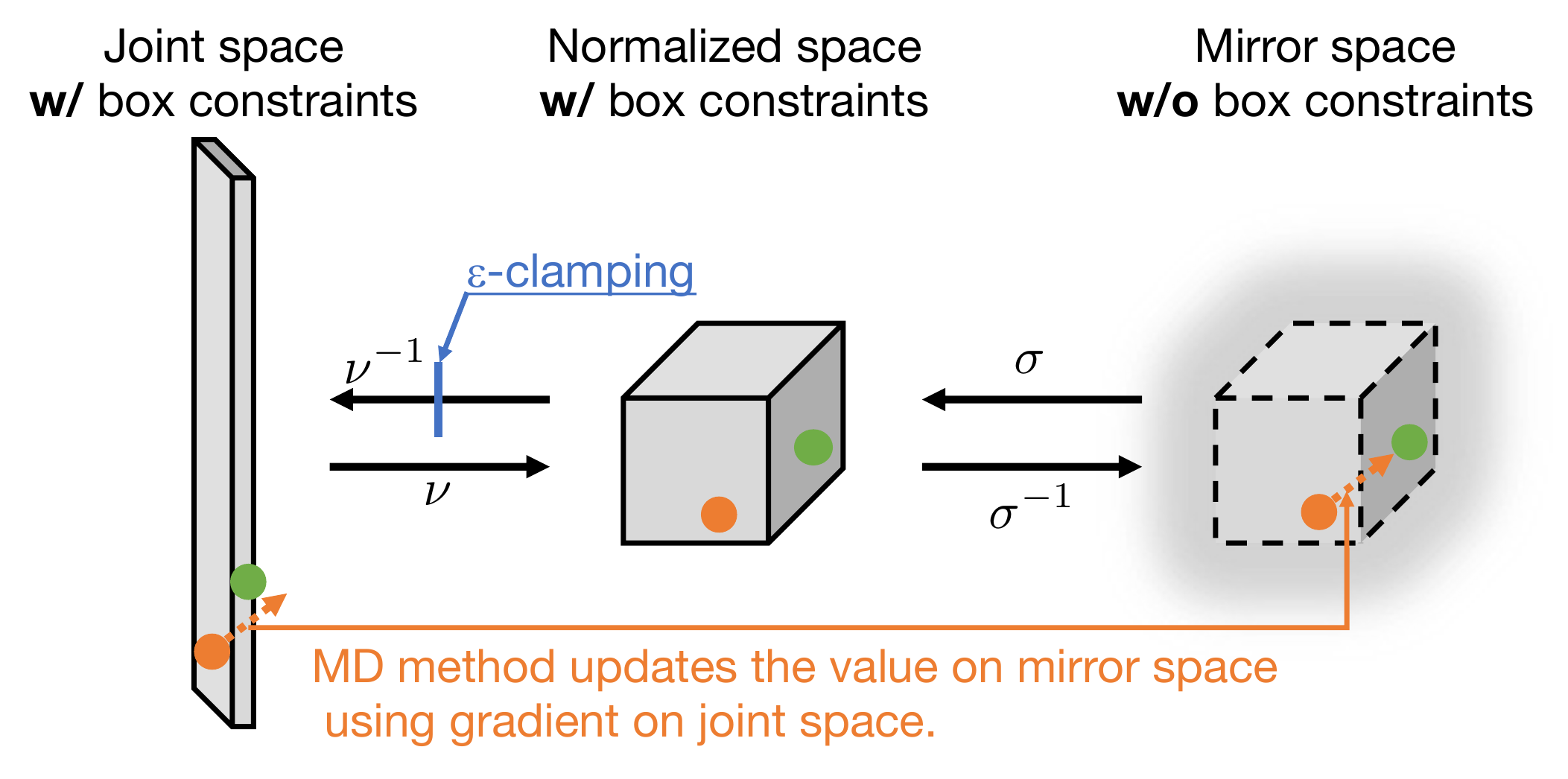}
    \caption{Conceptual scheme of MD-IK}
    \label{fig:concept_mdik}
\end{figure}

This paper therefore proposes the algorithm to integrate the conventional Jacobian-based IK with MD for joint constraints (box constraints as the first step of this study), so-called MD-IK (see Fig.~\ref{fig:concept_mdik}).
This integration allows MD-IK to solve the reference joints through minimizing a smooth objective function while considering the box-constrained joint space.
MD-IK is implemented with three design-related contributions as follows:
\begin{enumerate}
    \item Nonlinear mapping between the non-constrained mirror space and the box-constrained joint space is designed with mathematical derivation.
    \item A heuristic technique, called $\epsilon$-clamping, is introduced in order to escape from local optima caused nearly on the boundaries of constraints.
    \item An acceleration method~\citep{krichene2015accelerated} is modified with the assumption about the continuity of solutions at each time.
\end{enumerate}
Here, the second corresponds to the software-level margin.
In addition, as the third contribution, the interpretation as an MD method makes it possible to apply an acceleration method to it.
On top of that, an initialization that effectively utilizes continuity with past solutions is designed to increase the convergence speed.

For verification of MD-IK implemented on Pinocchio library~\citep{carpentier2019pinocchio}, a numerical tracking task is conducted with several types of robots.
Due to the benefit of the acceleration method, MD-IK acquires smaller tracking errors and smoother trajectories than the representative of Jacobian-based IK methods~\citep{sugihara2011solvability}.
Furthermore, in comparison with OSQP~\citep{stellato2020osqp} as the representative of QP solvers, MD-IK shows the same level of control performance with more smoothness and less computational cost.
In an additional walking task on a real-time dynamical simulator, OSQP often fails to long-term walking since it is adversely affected by the delay according to the time limit of optimization.
In contrast, MD-IK is able to continue walking stably and without such a delay even in scenarios where real-time control was severely required, thanks to the low computational cost suitable for holding the time limit.

\section{Preliminaries}

\subsection{Jacobian-based inverse kinematics}

Here, the conventional Jacobian-based IK~\citep{buss2004introduction,aristidou2018inverse} is briefly introduced.
The end effector pose as the combination of position $\boldsymbol{p}(\boldsymbol{q}) \in \mathbb{R}^3$ and orientation $Q(\boldsymbol{q})$ (unit quaternion) can be given as the functions of joint configurations $\boldsymbol{q} = [q_1, \ldots, q_N]^\top$ with $N$ active joints.
Given the reference pose as $\boldsymbol{p}^\mathrm{ref}$ and $Q^\mathrm{ref}$, the errors $\boldsymbol{e} \in \mathrm{R}^6$ between it and the current pose is also regarded as the function of $\boldsymbol{q}$ as follows:
\begin{align}
    \boldsymbol{e}(\boldsymbol{q}) = [
    \{\boldsymbol{p}^\mathrm{ref} - \boldsymbol{p}(\boldsymbol{q})\}^\top,
    f(Q^\mathrm{ref} * Q^{-1}(\boldsymbol{q}))^\top
    ]^\top
    \label{eq:ik_error}
\end{align}
where $f(\cdot) \in \mathbb{R}^3$ denotes the rotation vector representation.
The objective of IK is to find $\boldsymbol{q}^*$ as below.
\begin{align}
    \boldsymbol{q}^* = \arg \min_{\boldsymbol{q}} E(\boldsymbol{q})
    \text{, s.t. }
    E(\boldsymbol{q}) = \frac{1}{2} \boldsymbol{e}(\boldsymbol{q})^\top W \boldsymbol{e}(\boldsymbol{q})
    \label{eq:ik_problem}
\end{align}
where $W$ denotes the diagonal weight matrix.

To minimize this problem, the gradient descent method is one choice.
Specifically, the gradients $\boldsymbol{g}(\boldsymbol{q}) = \partial E / \partial \boldsymbol{q} \in \mathbb{R}^N$ can be derived as follows:
\begin{align}
    \boldsymbol{g}(\boldsymbol{q}) = - J^\top(\boldsymbol{q}) W \boldsymbol{e}(\boldsymbol{q})
    \label{eq:ik_jt}
\end{align}
where $J \in \mathbb{R}^{6 \times N}$ denotes Jacobian between the pose and the corresponding joint configurations.
Using $\boldsymbol{g}$, the reference joint configurations $\boldsymbol{q}^\mathrm{ref}$ are iteratively updated with the step size $\alpha > 0$.
\begin{align}
    \boldsymbol{q}^\mathrm{ref} &\gets \boldsymbol{q}^\mathrm{ref} - \alpha \boldsymbol{g}(\boldsymbol{q}^\mathrm{ref})
    \label{eq:ik_update}
\end{align}
where the initial $\boldsymbol{q}^\mathrm{ref}$ is often given as the observed joint configurations $\boldsymbol{q}^\mathrm{obs}$.
Until convergence (i.e. $E < \delta$ with $\delta$ threshold) or time limit, the above update is iterated.
This method is the Jacobian transpose (JT) method since the gradients in the method are computed with $J^\top$.

Note that if $J$ is assumed to be constant with respect to the variation range of $\boldsymbol{q}$, the above minimization problem can be attributed to QP using $J$.
Therefore, in recent years, QP solvers~\citep{ferreau2014qpoases,stellato2020osqp} have been increasingly used for IK.
However, although they are also iterative optimizations, the time required per iteration is longer than that of the Jacobian-based IKs, and there are cases where the optimization time exceeds the time limit or the optimization is not completed in time (see experimental results later).

To improve the convergence performance of the JT method, IK based on the Newton-Raphson method has been widely employed~\citep{sugihara2011solvability}, but the MD-IK proposed in this paper will extend the vanilla JT method.
Even so, the MD method has been proved to be equivalent to natural gradient method~\citep{raskutti2015information}, which can be viewed as a type of second-order gradient method~\citep{martens2020new}.
That is, the MD method may be interpreted as the second-order gradient method (like Newton-Raphson method), hence the sufficient convergence performance can be expected.

\subsection{Clamping for box-constrained joint space}

In practice, the joint configurations $\boldsymbol{q}$ are constrained mainly by upper and lower limits, $\boldsymbol{q}^\mathrm{low}$ and $\boldsymbol{q}^\mathrm{up}$ respectively ($\boldsymbol{q}^\mathrm{low} \leq \boldsymbol{q}^\mathrm{up}$), and maximum (angular) velocities $\dot{\boldsymbol{q}}^\mathrm{max}$ ($\dot{\boldsymbol{q}}^\mathrm{max} \geq 0$).
This paper, therefore, considers the following box constraints.
\begin{align}
    \underline{\boldsymbol{q}} \leq \boldsymbol{q} \leq \overline{\boldsymbol{q}}
    \label{eq:box_constraint}
\end{align}
where
\begin{align}
    \underline{\boldsymbol{q}} &= \max(\boldsymbol{q}^\mathrm{low}, \boldsymbol{q}^\mathrm{obs} - \dot{\boldsymbol{q}}^\mathrm{max} dt)
    \label{eq:box_inf} \\
    \overline{\boldsymbol{q}} &= \min(\boldsymbol{q}^\mathrm{up}, \boldsymbol{q}^\mathrm{obs} + \dot{\boldsymbol{q}}^\mathrm{max} dt)
    \label{eq:box_sup}
\end{align}
where $dt$ denotes the time step for control.

When $\boldsymbol{q}^\mathrm{ref}$ is updated by eq.~\eqref{eq:ik_update} with eq.~\eqref{eq:ik_jt}, it may violate this constraints since the gradients are in real space and the update law just subtracts them from $\boldsymbol{q}^\mathrm{ref}$.
In practical implementations to satisfy the constraints, the following clamping operation is conducted for $\boldsymbol{q}^\mathrm{ref}$ after each update iteration with eq.~\eqref{eq:ik_jt}.
\begin{align}
    \boldsymbol{q}^\mathrm{ref} \gets \max(\min(\boldsymbol{q}^\mathrm{ref}, \overline{\boldsymbol{q}}), \underline{\boldsymbol{q}})
    \label{eq:ik_clamp}
\end{align}

\subsection{Mirror descent method}

The non-smooth minimization problem by the clamping operation would cause several local minima, which would make IK numerically unstable~\citep{beeson2015trac}.
This paper therefore considers the box constraints in eq.~\eqref{eq:box_constraint} explicitly.
To this end, by focusing on the fact that the above Jacobian-based IK with clamping is regarded as the PG method, the MD method~\citep{beck2003mirror,srebro2011universality} is employed.
Its brief introduction is described below.

The MD method supposes the existence of mirror space, where the variables $\boldsymbol{q}$ in main space (i.e. constrained joint space in IK) can be mapped through nonlinear invertible mapping function as the mirror variables $\boldsymbol{\rho} = \psi(\boldsymbol{q})$ with the same dimension size as $\boldsymbol{q}$ and no constraint.
Although this mapping function is theoretically derived as the derivative of some Bregman divergences $D_\psi$, we can use the MD method with the arbitrary invertible function.
Given the minimization target $E(\boldsymbol{q})$, the update law of $\boldsymbol{q}$ is summarized as follows:
\begin{align}
    \boldsymbol{q} \gets \psi^{-1}\left (\psi(\boldsymbol{q}) - \alpha \frac{\partial E(\boldsymbol{q})}{\partial \boldsymbol{q}} \right )
    \label{eq:md_update}
\end{align}

As a remark, if $\partial E(\psi^{-1}(\boldsymbol{\rho}))/\partial \boldsymbol{\rho}$ is utilized instead of $\partial E(\boldsymbol{q})/\partial \boldsymbol{q}$, that method is consistent with the parameterization method.
That is, $\boldsymbol{\rho}$ is updated by the standard gradient descent method, and then, the updated value is mapped to $\boldsymbol{q}$ as the new value.
Indeed, the previous work~\citep{drexler2012joint} derives the gradients of $\boldsymbol{\rho}$ by multiplying Jacobian over $\boldsymbol{q}$ with $\partial \psi^{-1}(\boldsymbol{\rho}) / \boldsymbol{\rho}$, which would be less than $1$ and make convergence poor.
It is, therefore, faster to use $\partial E(\boldsymbol{q})/\partial \boldsymbol{q}$ as in the MD method.

\section{Proposal: MD-IK}

\subsection{Mapping between real joint and mirror spaces}

The proposed method, MD-IK, integrates the MD method with the JT-based IK in order to explicitly consider the box-constrained joint space.
For this integration, the invertible mapping function $\psi$ should be designed so that $\psi^{-1}$ maps real value on the mirror real space to one on the box-constrained joint space within $[\underline{\boldsymbol{q}}, \overline{\boldsymbol{q}}]$.

Such nonlinear mapping functions can be designed by using element-wise sigmoid function $\sigma$, such as logistic function and error function.
In general, however, the sigmoid function is a map to $[0, 1]$ (or $[-1, 1]$), and therefore, the denormalization function $\nu^{-1}$ is also required.
That is, the mapping function $\psi$ can be given as the following composite function of the inverted sigmoid function and the normalization function.
\begin{align}
    \psi &= \sigma^{-1} \circ \nu
    \label{eq:func_composite} \\
    \nu(\boldsymbol{q}) &= (\boldsymbol{q} - \underline{\boldsymbol{q}}) / (\overline{\boldsymbol{q}} - \underline{\boldsymbol{q}})
    \label{eq:func_normalize}
\end{align}
From the definition of $\nu$, we can easily compute $\nu^{-1}$ as the denormalization function, $\boldsymbol{q} = (\overline{\boldsymbol{q}} - \underline{\boldsymbol{q}}) \nu(\boldsymbol{q}) + \underline{\boldsymbol{q}}$.
Using the above functions, the general update law of MD-IK is derived as follows:
\begin{align}
    \boldsymbol{q}^\mathrm{ref} &\gets \psi^{-1}\left (\psi(\boldsymbol{q}^\mathrm{ref}) - \alpha \boldsymbol{g}(\boldsymbol{q}^\mathrm{ref}) \right )
    \nonumber \\
    &= (\overline{\boldsymbol{q}} - \underline{\boldsymbol{q}})
    \sigma \left (\sigma^{-1} \left ( \frac{\boldsymbol{q}^\mathrm{ref} - \underline{\boldsymbol{q}}}{\overline{\boldsymbol{q}} - \underline{\boldsymbol{q}}} \right )
    - \alpha \boldsymbol{g}(\boldsymbol{q}^\mathrm{ref}) \right )
    + \underline{\boldsymbol{q}}
    \label{eq:md_ik_general}
\end{align}
Note that if $\underline{\boldsymbol{q}} = \overline{\boldsymbol{q}}$, the above computation should be ignored and forcibly set $\boldsymbol{q}^\mathrm{ref} = \overline{\boldsymbol{q}}$.

\begin{figure}[tb]
    \centering
    \includegraphics[keepaspectratio=true,width=0.96\linewidth]{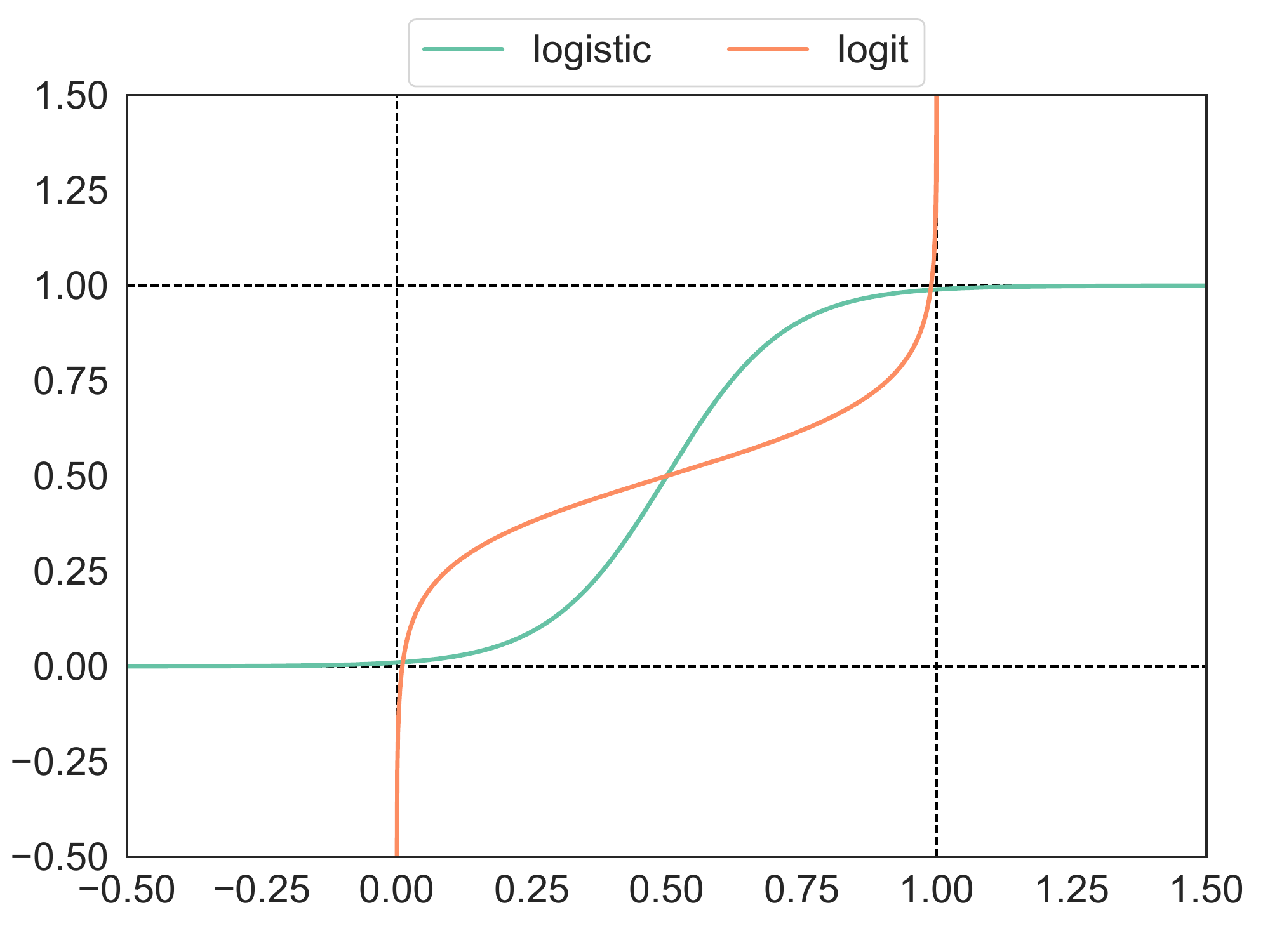}
    \caption{Example of sigmoid function $\sigma$ and its inverse $\sigma^{-1}$}
    \label{fig:func_mirror}
\end{figure}

Here, heuristic conditions for $\sigma$ are given as below.
Specifically, since only the normalized value (minus the gradient) is fed into $\sigma$, it is desired for $\sigma$ to output roughly $(0, 1)$ in that range to fully use the entire range of $\sigma$.
This requirement leads to the following conditions with a threshold $\epsilon \ll 1$.
\begin{align}
    \sigma(0) = \epsilon
    , \
    \sigma(1) = 1 - \epsilon
    \label{eq:sigmoid_cond}
\end{align}
$\sigma$ designed under the above conditions and its inverse $\sigma^{-1}$ are illustrated in Fig.~\ref{fig:func_mirror}.
As can be seen, almost the entire function shapes are contained within $[0, 1]$.

\subsection{$\epsilon$-clamping as margin in software level}

The MD method with the above sigmoid function yields the smooth update without clamping.
On the other hand, it makes the update difficult to escape nearly on the boundaries.
To mitigate this local optima problem, a heuristic technique, called $\epsilon$-clamping, is implemented.

Specifically, the clamping with $\epsilon$ is applied after eq.~\eqref{eq:md_ik_general}.
\begin{align}
    \boldsymbol{q}^\mathrm{ref} \gets \nu^{-1}(\max(\min(\nu(\boldsymbol{q}^\mathrm{ref}), 1 - \epsilon), \epsilon))
    \label{eq:ik_clamp_eps}
\end{align}
This clamping can be regarded as a margin at the software level to avoid boundary values that are not desirable for commands in terms of practical joint control performance.
For example, the maximum velocity is a catalog specification and would not be performed, and the hardware boundaries would cause mechanical collisions.
In addition, in terms of smoothness, the effect of this clamping is insignificant since the gradients on the clamped range converges to almost zero when $\sigma$ is designed according to eq.~\eqref{eq:sigmoid_cond}.
Note that the box constraints with $\epsilon$-clamping is slightly shrunk as
$[(1-\epsilon) \underline{\boldsymbol{q}} + \epsilon \overline{\boldsymbol{q}}, (1-\epsilon) \overline{\boldsymbol{q}} - \epsilon \underline{\boldsymbol{q}}]$.

\subsection{Smoothly accelerated MD-IK}

Finally, in order to further improve the convergence of MD-IK, a method corresponding to Nesterov acceleration method in the MD method~\citep{krichene2015accelerated}, which combines the MD and PG methods appropriately, is applied.
In IK, due to the very limited computation time, fast convergence (by the PG method) in the early stage of iteration and smoothness of the solution (by the MD method) in the late stage of iteration are considered to be required.
In addition, to suppress memory cost, it would be preferable to use the PG method to compensate for the iteration results of the MD method, rather than using the two independently.

From these implementation requirements, the following update rule is applied after updating $\boldsymbol{q}^\mathrm{ref}$ by the MD method with $\epsilon$-clamping in eqs.~\eqref{eq:md_ik_general}--\eqref{eq:ik_clamp_eps}.
\begin{align}
    \boldsymbol{z} &\gets \max(\min(\boldsymbol{z} - \alpha_z \boldsymbol{g}(\boldsymbol{q}^\mathrm{ref}), \overline{\boldsymbol{q}}), \underline{\boldsymbol{q}})
    \label{eq:md_ik_accel_pg} \\
    \boldsymbol{q}^\mathrm{ref} &\gets \beta \boldsymbol{z} + (1 - \beta) \boldsymbol{q}^\mathrm{ref}
    \label{eq:md_ik_accel_ip}
\end{align}
where $\boldsymbol{z}$ denotes the variables updated by the PG method with its own step size $\alpha_z = (k \alpha)/(r \gamma)$ with $k$ the number of iterations, $r$ the smoothness of switching between the PG and MD methods, and $\gamma$ the hyperparameter to adjust the step size for the PG method.
$\beta = (1 + k/r)^{-1}$ determines the ratio when interpolating $\boldsymbol{z}$ and $\boldsymbol{q}^\mathrm{ref}$.

Basically, the initial $\boldsymbol{z}$ is given as $\boldsymbol{q}^\mathrm{obs}$, which is the initial $\boldsymbol{q}^\mathrm{ref}$, and $k$ is reset to be $1$.
In IK, however, the continuity of the solution can be assumed since the target trajectory is constrained by the kinodynamic performance.
Under this assumption, it is easily expected that the previously optimized $\boldsymbol{z}$ can be reused, and $k$ does not need to be completely reset.
Therefore, instead of treating IK at each time independently, the following smooth reset is introduced to make effective use of past IK results.
\begin{align}
    \boldsymbol{z} \gets \max(\min(\eta \boldsymbol{z} + (1 - \eta) \boldsymbol{q}^\mathrm{obs}, \overline{\boldsymbol{q}}), \underline{\boldsymbol{q}})
    \text{, }
    k \gets \eta k
    \label{eq:md_ik_smooth}
\end{align}
where $\eta \in [0, 1]$ denotes the smoothness of resetting, and for the first time only, $\boldsymbol{z} = \boldsymbol{q}^\mathrm{obs}$ and $k=1$ as usual.
Note that the box constraint changes at each time, so the clamping process is performed just in case.

In summary, the pseudo code of MD-IK is described in Alg.~\ref{alg:proposal}.
Although it is possible to update $J$ at each iteration in the Jacobian-based IK, we assume that the variation of $J$ is small and fix it at the initial $J$ as in QP, in order to reduce computational cost.
In addition, $\zeta \in (0, 1)$ is introduced to make the time limit shorter than the control period $dt$.
The larger $\zeta$, the more accurate the convergence can be, but the time delay until $\boldsymbol{q}^\mathrm{ref}$ is commanded to the robot will increase, which can destabilize the robot motion.

\begin{algorithm}[tb]
    \caption{MD-IK}
    \label{alg:proposal}
    \begin{algorithmic}[1]
        \algnewcommand{\Break}[1]{\State\algorithmicif\ #1\ \algorithmicthen\ \textbf{break}}
        \State{Get $\boldsymbol{q}^\mathrm{low}$, $\boldsymbol{q}^\mathrm{up}$, and $\dot{\boldsymbol{q}}^\mathrm{max}$ from model}
        \State{Give $W$, $\alpha$, $\delta$, $dt$, $\zeta$, $\epsilon$, $r$, $\gamma$, and $\eta$}
        \State{Set $\boldsymbol{p}^\mathrm{ref}$ and $Q^\mathrm{ref}$}
        \State{Observe $\boldsymbol{q}^\mathrm{obs}$, and set $\boldsymbol{q}^\mathrm{ref} = \boldsymbol{q}^\mathrm{obs}$}
        \State{Set $\underline{\boldsymbol{q}}$ and $\overline{\boldsymbol{q}}$ by eqs.~\eqref{eq:box_inf} and~\eqref{eq:box_sup}, and compute $J(\boldsymbol{q}^\mathrm{ref})$}
        \State{Set $\boldsymbol{z}$ and $k$ by eq.~\eqref{eq:md_ik_smooth} and $t_s$ as the current time $t$}
        \While{True}
            \State{Compute $\boldsymbol{e}(\boldsymbol{q}^\mathrm{ref})$ and $E(\boldsymbol{q}^\mathrm{ref})$ by eqs.~\eqref{eq:ik_error} and~\eqref{eq:ik_problem}}
            \Break{$(E(\boldsymbol{q}^\mathrm{ref}) < \delta)$ or $(t - ts > \zeta dt)$}
            \State{Compute $\boldsymbol{g}(\boldsymbol{q}^\mathrm{ref})$ by eq.~\eqref{eq:ik_jt}}
            \State{Update $\boldsymbol{q}^\mathrm{ref}$ by the MD method in eq.~\eqref{eq:md_ik_general}}
            \State{Clamp $\boldsymbol{q}^\mathrm{ref}$ by the $\epsilon$-clamping in eq.~\eqref{eq:ik_clamp_eps}}
            \State{Update $\boldsymbol{z}$ and $\boldsymbol{q}^\mathrm{ref}$ by eqs.~\eqref{eq:md_ik_accel_pg} and~\eqref{eq:md_ik_accel_ip}}
            \State{$k += 1$}
        \EndWhile
        \State{Return $\boldsymbol{q}^\mathrm{ref}$}
    \end{algorithmic}
\end{algorithm}

\section{Experiment}

\subsection{Configurations}

\begin{table}[tb]
    \caption{Hyperparameters}
    \label{tab:method_config}
    \centering
    {
    \begin{tabular}{ccc}
        \hline\hline
        Symbol & Meaning & Value \\
        \hline
        $W$ & Weight matrix for minimization target & $I$ \\
        $\alpha$ & Step size for update & $1$ \\
        $\delta$ & Threshold for convergence & $1 \times 10^{-10}$ \\
        $\epsilon$ & Threshold for $\epsilon$-clamping & $1 \times 10^{-2}$ \\
        $r$ & Smoothness of switching PG and MD & $5$ \\
        $\gamma$ & Reduction of PG step size & $2$ \\
        $\eta$ & Ratio of smooth reset & $0.5$ \\
        $\lambda$ & Damping factor & $1 \times 10^{-3}$ \\
        \hline\hline
    \end{tabular}
    }
\end{table}

To compute the Jacobian matrix from robot models, a python binding of Pinocchio library~\citep{carpentier2019pinocchio}, which contains rigid body algorithms mainly for humanoid robots, was employed.
The core of MD-IK was implemented by NumPy with JIT compile.
For comparison, IK with Levenberg-Marquardt~\citep{sugihara2011solvability} was implemented as LM based on this code (its minimum damping factor $\lambda$ is set to be $1 \times 10^{-3}$).
Another comparison, OSQP~\citep{stellato2020osqp}, was also implemented using its python binding with the default configurations.
Note that, for numerical stability, $\lambda I$ was added to $J^\top W J$ when making a QP.
Although qpOASES~\citep{ferreau2014qpoases} was also tested, its results were omitted in this paper because it mostly generated unnatural behaviors within the time limit.

The hyperparameters specified are given in Table~\ref{tab:method_config}.
Note that all the source codes used were uploaded on GitHub:
\url{https://github.com/kbys-t/mdik}.

\begin{figure*}[tb]
    \begin{minipage}[b]{0.32\linewidth}
        \centering
        \includegraphics[keepaspectratio=true,width=\linewidth]{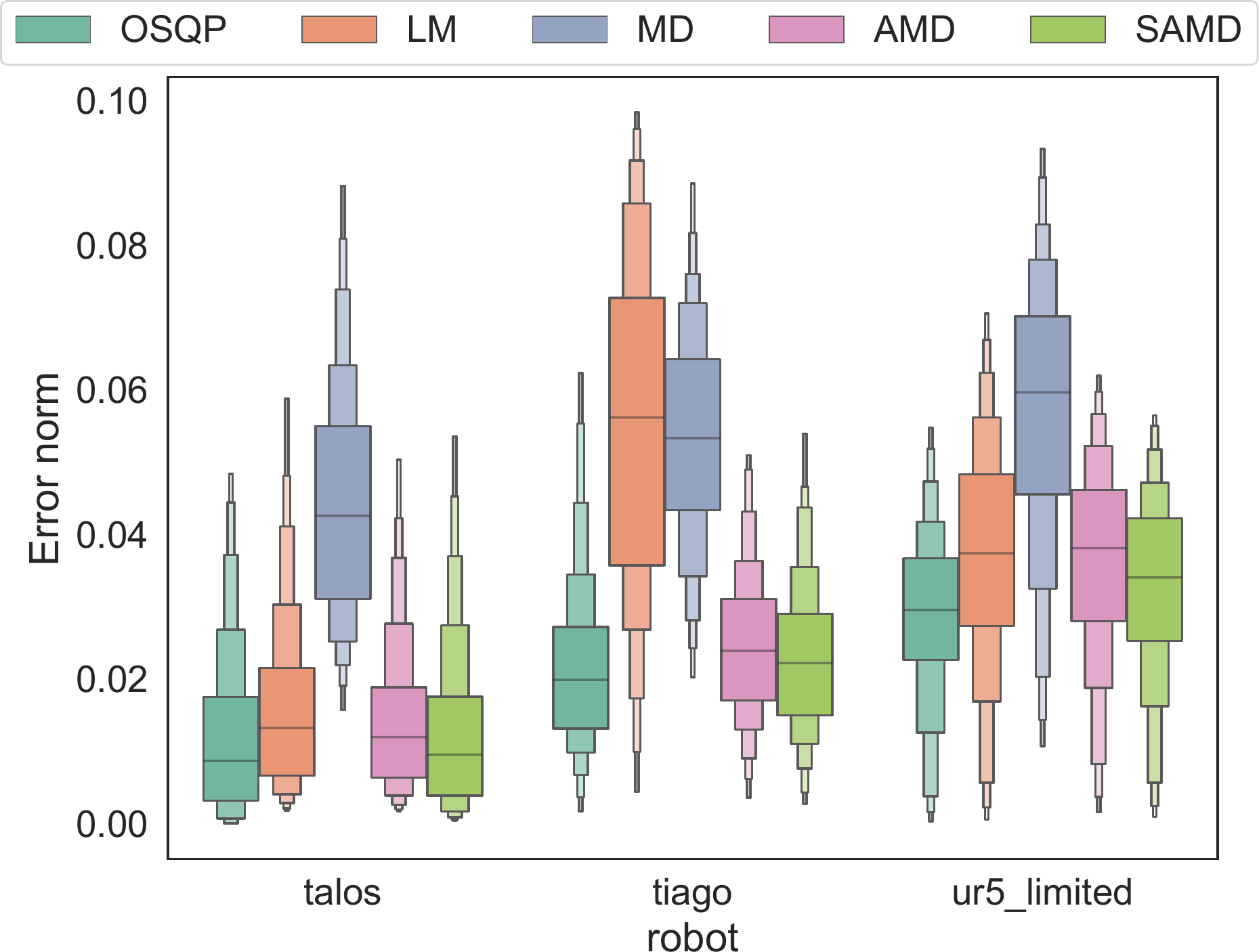}
        {(a) Error norm}
    \end{minipage}
    \begin{minipage}[b]{0.32\linewidth}
        \centering
        \includegraphics[keepaspectratio=true,width=\linewidth]{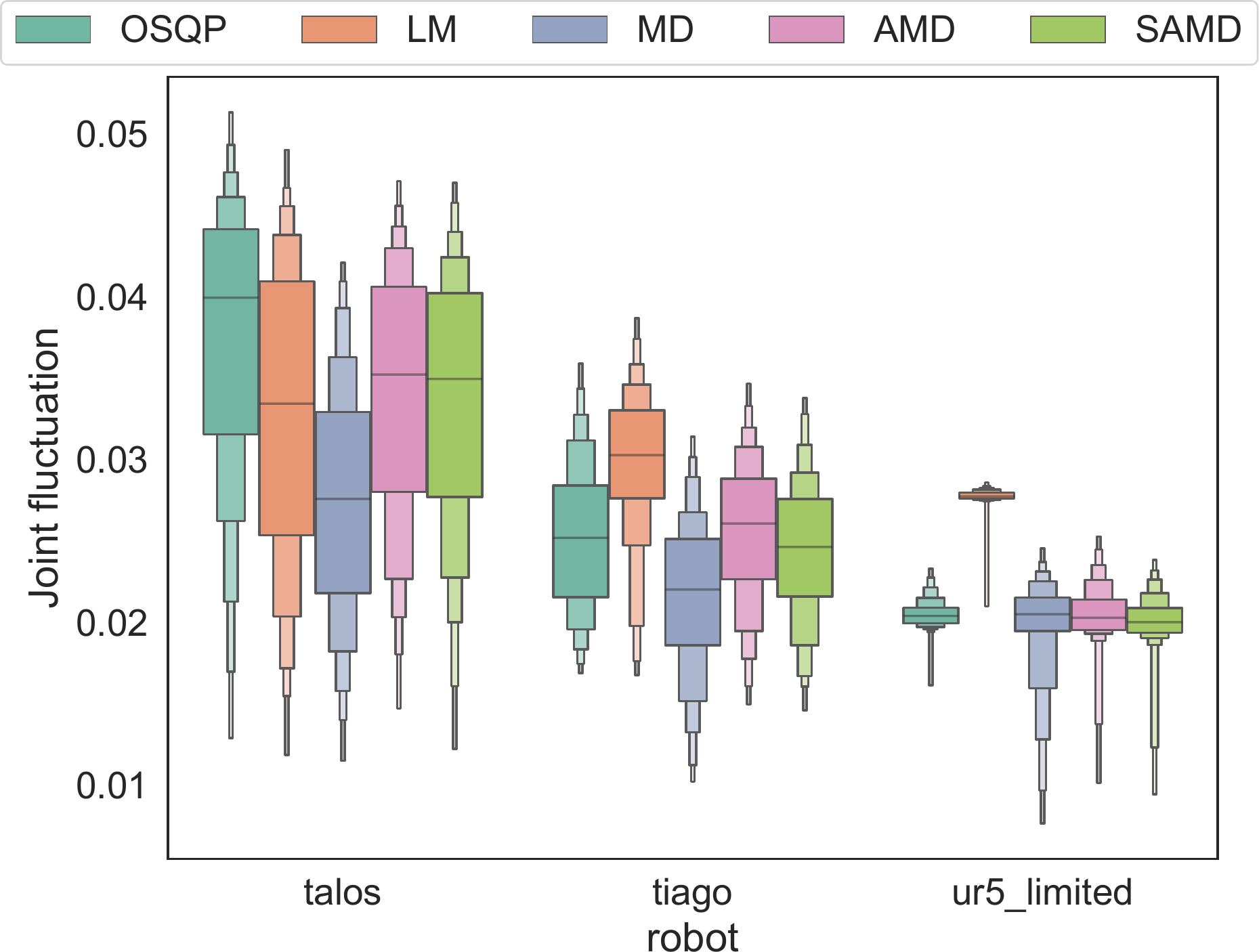}
        {(b) Joint fluctuation}
    \end{minipage}
    \begin{minipage}[b]{0.32\linewidth}
        \centering
        \includegraphics[keepaspectratio=true,width=\linewidth]{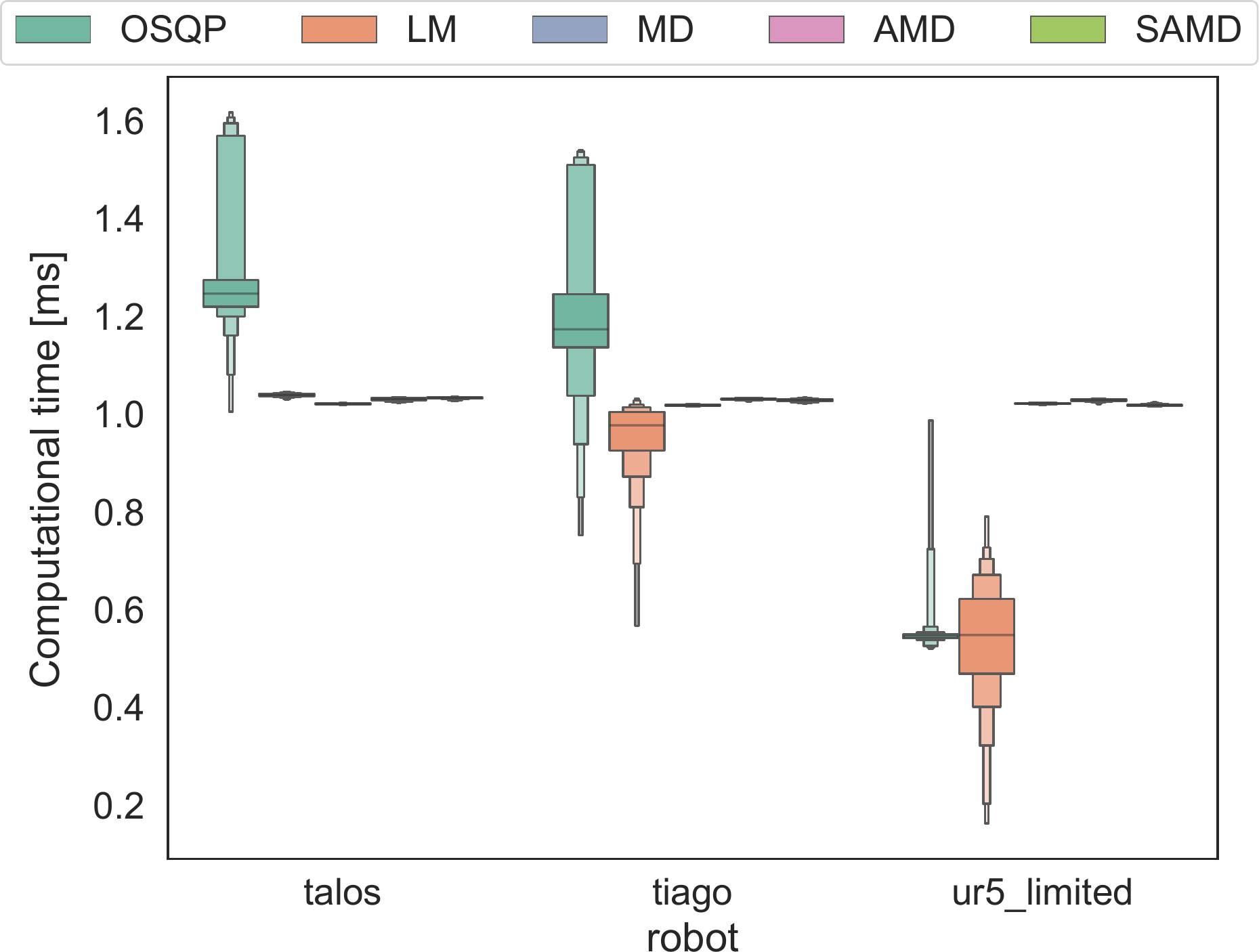}
        {(c) Computational time}
    \end{minipage}
    \caption{Tracking results}
    \label{fig:tracking}
\end{figure*}

\subsection{Tracking task}

\subsubsection{Problem statement}

In this task, each component of the reference pose of the robot's end effector is moved within the given area by a sinusoidal wave (0.5 Hz at maximum, and uniformly randomly given), and the robot tracks it during 12.5 sec.
Given $dt = 5 \times 10^{-3}$ and $\zeta = 0.2$, the IK computation time is within 1~ms.
500 trials for the tracking task are performed under each condition with different reference pose or trajectory, and the performance of each condition is statistically evaluated while excluding trials given trajectories that are clearly difficult to follow with any IKs.
Note that during trials, only box-constraints on joint space are considered and dynamical constraints and self collisions are ignored for simplicity.

Three different robot models are used for this verification, taking into account the effect of structural differences in the robots.
Specifically, TALOS~\citep{stasse2017talos} is employed as a full-sized humanoid robot with redundant joints.
Its left hand is set as the end effector.
As a redundant robot with prismatic joint, TIAGo~\citep{pages2016tiago} is employed.
Its torso is lifted up/down by a linear actuator.
UR5~\citep{kebria2016kinematic} is finally employed as a six degree-of-freedom robot, IK of which is actually able to be computed analytically.

\subsubsection{Results}

The experimental results are summarized in Fig.~\ref{fig:tracking}, in which (a) and (b) indicate the control performance and the smoothness of the trajectory (smaller is better for both), respectively.
In the legend, OSQP and LM are the conventional methods for comparison; and others are MD without acceleration, AMD with the acceleration method by eqs.~\eqref{eq:md_ik_accel_pg} and~\eqref{eq:md_ik_accel_ip}, and SAMD with eq.~\eqref{eq:md_ik_accel_ip} for the smooth reset.

First, it is noticeable that LM and MD caused poor control performance, although MD has high smoothness.
Second, while OSQP has high control performance, it exceeded the specified computation time (i.e. 1~ms).
While a margin in computation time would allow the implementation to maintain real-time control with OSQP, this computation delay could adversely affect the motion (see the next task).
In contrast, AMD and SAMD achieved approximately the same level of control performance as OSQP and higher smoothness than OSQP, while keeping the time limit.
In addition, SAMD slightly improved performance in all conditions from AMD.

\subsection{Walking task}

\subsubsection{Problem statement}

In this task, a humanoid robot, Atlas~\citep{nelson2019petman}, tries to walk on a real-time dynamics simulator~\citep{coumans2016pybullet}.
The walking trajectory is generated as a box step based on the literature~\citep{kobayashi2022whole}.
Five end effectors (i.e. both legs, both hands, and a torso) are registered for solving IK: namely, Jacobians for the respective end effectors are concatenated as $J \in \mathbb{R}^{30 \times N}$, and then IK is solved with it.
Note that the relative reference poses for both hands and the torso are fixed only to stabilize motion.
Given $dt = \{1/480, 1/720\}$ and $\zeta = 0.36$, the IK computation times are within \{0.75, 0.5\}~ms, respectively.
By making the walk controller open-loop, the real-time control performance of IK is key to the continuation of the box step (30 steps in 5 rounds at maximum).

\subsubsection{Results}

\begin{figure}[tb]
    \centering
    \includegraphics[keepaspectratio=true,width=0.96\linewidth]{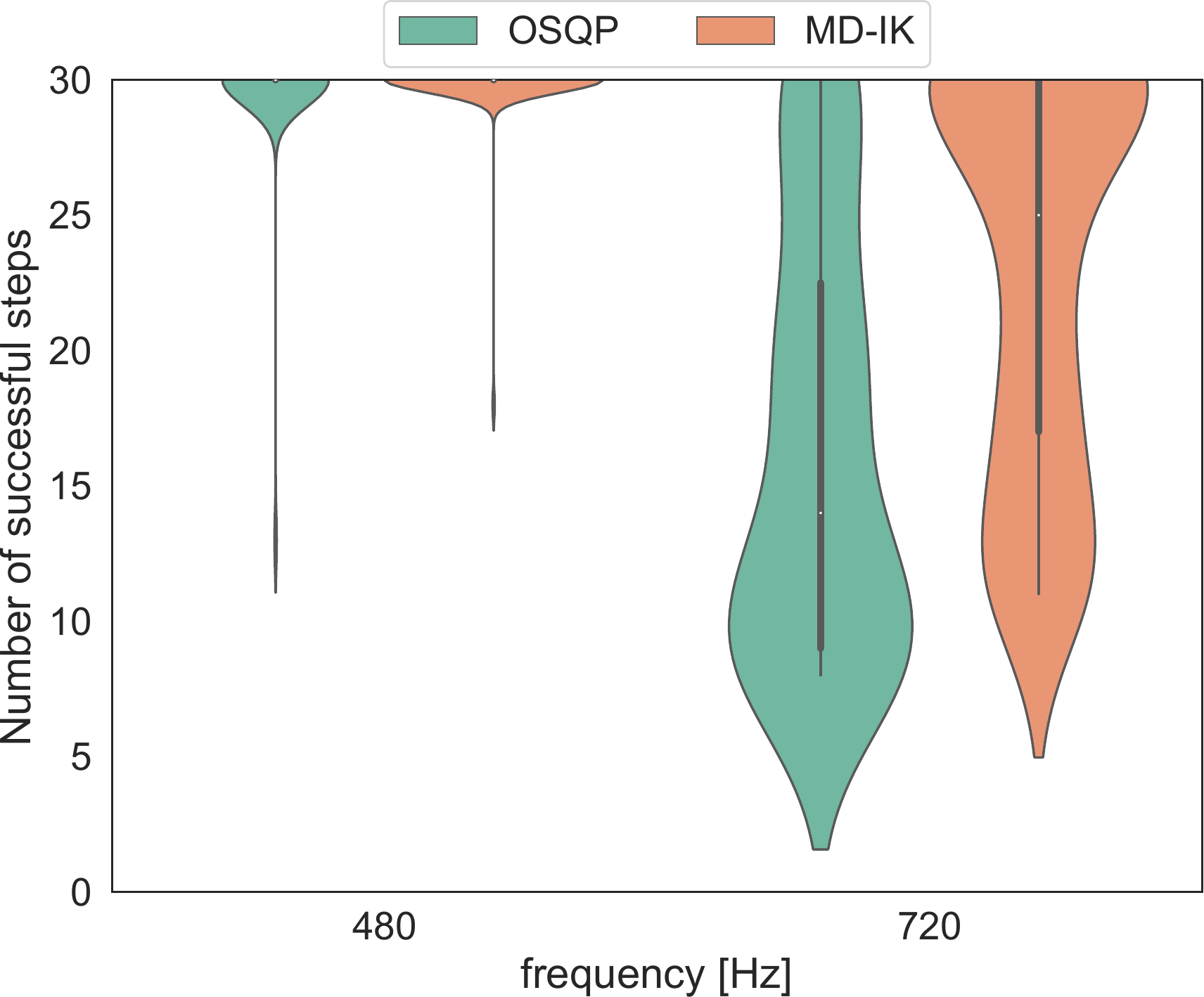}
    \caption{Walking results on the number of successful steps}
    \label{fig:walk_step}
\end{figure}

The experimental results are summarized in Fig.~\ref{fig:tracking}, which shows the number of successful steps in 100 trials.
Here, MD-IK denotes the proposed method described in Alg.~\ref{alg:proposal}.
Only OSQP was compared due to its high performance.

The results indicated that MD-IK achieved more stable locomotion along the reference walking trajectory.
The difference is particularly noticeable when the control period is 720~Hz.
Indeed, OSQP was unable to solve IK within the time limit, and the delay in updating the reference joint angles due to the excess time contributed significantly to the instability of the locomotion.
In contrast, MD-IK was able to solve IK with sufficient accuracy within the time limit, and thus was able to continue walking without the effect of the delay.
Note that examples of this experiment were uploaded to YouTube \url{https://youtu.be/GFO6trmgbn8}.

\section{Conclusion}

This paper proposed a new Jacobian-based IK solver explicitly considering box-constrained joint space based on the MD method, so-called MD-IK.
Specifically, the box-constrained joint space is mapped to the no-constrained and invertible mirror space, and Jacobian-based gradient is applied to update the no-constrained mirror variables, which is remapped to the box-constrained joint references.
In addition, to escape local optima nearly on the boundaries of constraints, $\epsilon$-clamping is heuristically implemented as margin at software level.
The Nesterov-based acceleration method is integrated with MD-IK while smoothly utilizing the past IK results.
As a result, MD-IK can achieved the same level of control performance as OSQP, one of the current mainstream IK solvers, with more smoothness and less computational cost.
This is practically important, and in fact, MD-IK actually outperformed OSQP in real-time walking simulation.

Although this paper derived a method limited to box constraints for the independent joints, but if an appropriate invertible mapping can be constructed, it should be possible to handle coupled constraints that can take self-collision etc. into account.
In the future, therefore, we will improve the generality of MD-IK so that it can be applied to such constraints.

%
%
\bibliographystyle{ifacconf}
{
\bibliography{biblio}
}


\end{document}